%% file: paper.tex
\documentclass[runningheads]{llncs}
\pdfoutput=1

\usepackage[utf8]{inputenc}
\usepackage{amsmath,amssymb}
\usepackage{times,helvet,courier}
\usepackage{url}
\usepackage{listings}
\usepackage[FIGTOPCAP,TABTOPCAP,tight]{subfigure}
\usepackage{pgf}
\usepackage{tikz}
\usetikzlibrary{arrows,automata,shapes.multipart,matrix}
\usetikzlibrary[positioning]

\input{macro} 
\newcommand{\EXT}[2]{#2}% {#1} % {#2\marginpar{REDUCTION!}} %{#1\marginpar{EXTENSION!}}
\title{Answer Set Programming for Stream Reasoning\thanks{%
  This paper complements a short KR'12 paper~\cite{gegrkaobsasc12b};
  an extended draft~\cite{gegrkaobsasc12a} is available at~\cite{oclingo}.}}

\author{
  M.~Gebser\inst{1}
  \and
  T.~Grote\inst{1}
  \and
  R.~Kaminski\inst{1}
  \and
  P.~Obermeier\inst{2}
  \and
  O.~Sabuncu\inst{1}
  \and
  T.~Schaub\inst{1}\thanks{Affiliated with Simon Fraser University, Canada, and Griffith University, Australia.}}
\institute{Universit\"at Potsdam, Germany \and DERI Galway, Ireland}

\titlerunning{Answer Set Programming for Stream Reasoning}
\authorrunning{M.~Gebser \emph{et al}\/.}

\begin{document}

\setcounter{page}{115}

\maketitle

\input{abstract}

\input{introduction}

\input{background}
\input{modeling}
\input{access}
\input{traffic}

\input{onlinejobs}
\input{discussion}
\input{acknowledgments}
% \newpage
% \nocite{gegrkaobsasc12b}
%\bibliographystyle{splncs}
%\bibliography{lit,akku,procs}

\input{bbl}

\end{document}

%% file: macro.tex
\newcommand{\clasp}{\textit{clasp}}

\newcommand{\clingo}{\textit{clingo}}

\newcommand{\gringo}{\textit{gringo}}

\newcommand{\oclingo}{\textit{oclingo}}

\newcommand{\accessBase}{\lstinline{accessB.lp}}
\newcommand{\accessCumulative}{\lstinline{accessC.lp}}

\newcommand{\jobscheduleBase}{\lstinline{jobsB.lp}}
\newcommand{\overtakingBase}{\lstinline{overtakingB.lp}}

\newcommand{\lackgap}{}%\hspace{-2mm}}
\newcommand{\code}[1]{\lstinline[basicstyle=\tt\small]{#1}}
\newcommand{\breakeq}{\\[-1.5pt]}
\newcommand{\subfigsqueeze}{\vspace{-2.5pt}}

%% file: abstract.tex
\begin{abstract}
The advance of Internet and Sensor technology has brought about new 
challenges evoked by the emergence of continuous data streams.
Beyond rapid data processing, application areas like
ambient assisted living, robotics, or dynamic scheduling
involve complex reasoning tasks.
We address such scenarios
and elaborate upon approaches to knowledge-intense 
stream reasoning, 
based on Answer Set Programming (ASP).
While traditional ASP methods are devised for singular problem solving,
we develop new techniques to formulate and process
problems dealing with emerging as well as expiring data in a seamless way. 
% We thus provide novel
% approaches for modeling and reasoning with continuous data.
\end{abstract}

%%% Local Variables: 
%%% mode: latex
%%% TeX-master: "paper"
%%% End: 

%% file: introduction.tex
\section{Introduction}\label{sec:introduction}

The advance of Internet and Sensor technology has brought about new
challenges evoked by the emergence of continuous data streams, like
web logs, mobile locations, or online measurements. % traffic data.
While existing data stream management systems~\cite{golozs10a} allow for 
high-throughput stream processing,
they lack complex reasoning capacities~\cite{vacevafe09a}.
We address this shortcoming and elaborate upon approaches to knowledge-intense
stream reasoning,
based on Answer Set Programming (ASP;~\cite{baral02a})
as a prime tool for Knowledge Representation and Reasoning (KRR;~\cite{KRHandbook}).
The emphasis thus shifts from rapid data processing towards complex reasoning,
as % needed for instance in 
required in application areas like
ambient assisted living, robotics, or dynamic scheduling.

% However,
% the sheer amount and continuous flow of information produced by data streams
% precludes the direct application of ASP,
% simply because it is designed for singular reasoning from all available information.
% Unlike this, 
In contrast to traditional ASP methods, which are devised for singular problem solving,
{\em ``stream reasoning, instead, restricts processing to a certain window of
concern, focusing on a subset of recent statements in the stream, while ignoring
previous statements''}~\cite{babrcedegr10a}.
To accommodate this in ASP,
we develop new techniques % that allow us to formulate problem encodings 
to formulate and process problems
dealing with emerging as well as expiring data in a seamless way.
Our modeling approaches rely on the novel concept of
time-decaying logic programs~\cite{gegrkaobsasc12b}, where
logic program parts are associated with life spans to steer
their emergence as well as expiration upon continuous reasoning.
Time-decaying logic programs are implemented as a recent extension
of the reactive ASP system \oclingo~\cite{gegrkasc11a},
using the ASP grounder \gringo~\cite{gekakosc11a} for the recurrent
composition of a static ``offline'' encoding with
dynamic ``online'' data into queries to the ASP solver \clasp~\cite{gekanesc07a}.

While \oclingo\ makes powerful ASP technology accessible for stream reasoning,
its continuous query formulation and processing impose particular modeling challenges.
First, re-grounding parts of an encoding wrt.\ new data shall be as economical as possible.
Second, traditional modeling techniques, eg.\ frame axioms~\cite{lifschitz02a},
need to be reconsidered in view of the expiration of obsolete program parts.
Third, the re-use of propositional atoms and rules shall be extensive
to benefit from conflict-driven learning (cf.~\cite{SATHandbook}).
We here tackle such issues in % the context of 
continuous reasoning 
over data from a sliding window~\cite{golozs10a}.
In a nutshell, we propose to % statically 
encode knowledge about any potential window contents offline,
so that dynamic logic program parts can concentrate on activating readily available
rules (via transient online data).
% For one, this reduces re-grounding efforts to a minimum because it boils down
% to matching new data to a corresponding internal representation.
% For another, the static representation of all possible window contents
% maximizes re-use at the propositional level, thus enabling conflict-driven
% learning to gather problem information over successive queries.
Moreover, we show how default conclusions, eg.\ expressed by frame axioms,
can be faithfully combined with % emerging as well as expiring sliding window 
transient data.

After providing the necessary background,
we demonstrate % our % proposed 
modeling approaches on three toy domains.
The % basic ideas 
underlying principles
are, however, of general applicability and thus establish
universal patterns for ASP-based reasoning over sliding windows.\footnote{%
For formal details on time-decaying logic programs and a discussion of related work
in stream processing, we refer the interested reader to~\cite{gegrkaobsasc12b,gegrkaobsasc12a}.}

%% file: background.tex
\section{Background}\label{sec:background}

We presuppose familiarity with (traditional) ASP input languages
(cf.~\cite{potasscoManual,lparseManual}) and 
(extended) logic programs (cf.~\cite{siniso02a,baral02a}).
They provide the basis of 
\emph{incremental logic programs}~\cite{gekakaosscth08a},
where additional keywords,
``\code{\#base},''
``\code{\#cumulative},''
and
``\code{\#volatile},''
allow for partitioning rules into a static, an accumulating, and
a transient program part.
The latter two usually refer to some constant~\code{t}, standing for 
a step number.
In fact, when gradually increasing the step number, starting from~$1$,
ground instances of rules in a \code{\#cumulative} block are successively 
generated and joined with ground rules from previous steps,
whereas a \code{\#volatile} block contributes instances of its rules
for the current step only.
Unlike accumulating and transient program parts, which are re-processed at
each incremental step,
the static part indicated by \code{\#base} is instantiated just once, initially,
so that its rules correspond to a ``$0$th'' \code{\#cumulative} block.

Application areas of incremental logic programs
include planning (cf.~\cite{naghtr04a}) and finite model finding (cf.~\cite{gesasc11a}).
For instance, (a variant of) the well-known Yale shooting problem (cf.~\cite{baker89a}) can be modeled by an
incremental logic program as follows:
\begin{lstlisting}[numbers=none,basicstyle=\ttfamily\footnotesize,belowskip=0mm,aboveskip=0mm]
  #base.
  { loaded }.
  live(0).
  #cumulative t.
  { shoot(t+1) }.
    ab(t) :- shoot(t), loaded.
  live(t) :- live(t-1), not ab(t).
  #volatile t.
  :- live(t).
  :- shoot(t+1).
\end{lstlisting}
The first answer set,
containing % the atoms 
\code{loaded}, 
\code{live(0)},
\code{live(1)},
\code{shoot(2)}, and
\code{ab(2)},
is generated from the following ground rules at step~$2$:
\begin{lstlisting}[numbers=none,basicstyle=\ttfamily\footnotesize,belowskip=0mm,aboveskip=0mm]
  % #base.
  { loaded }.
  live(0).
  % #cumulative 1.                % #cumulative 2.
  { shoot(2) }.                   { shoot(3) }.
    ab(1) :- shoot(1), loaded.      ab(2) :- shoot(2), loaded.
  live(1) :- live(0), not ab(1).  live(2) :- live(1), not ab(2).
  % #volatile 2.
  :- live(2).
  :- shoot(3).
\end{lstlisting}
Observe that the static \code{#base} part is augmented with rules from
the \code{\#cumulative} block for step~$1$ and~$2$,
whereas \code{\#volatile} rules are included for step~$2$ only.

The \emph{reactive} ASP system \oclingo\ extends offline incremental
logic programs by functionalities to incorporate external online information
from a controller, also distinguishing accumulating and transient external inputs.
For example,
consider a character stream over alphabet
$\{a,b\}$ along with the task of continuously checking whether a stream prefix at
hand matches regular expression $(a|b)^*aa$.
To provide the stream prefix $aab$,
the controller component of \oclingo\ can successively pass facts as follows:
\begin{lstlisting}[numbers=none,basicstyle=\ttfamily\footnotesize,belowskip=0mm,aboveskip=0mm]
  #step 1. read(a,1).
  #step 2. read(a,2).
  #step 3. read(b,3).
\end{lstlisting}
The number~$i$ in ``\code{#step\ }$i$\code{.}'' directives informs 
\oclingo\ about the minimum step number up to which an underlying
offline encoding must be instantiated in order to incorporate
external information (given after a \code{#step} directive)
in a meaningful way.
In fact, the following offline encoding
builds on the assumption that values~$i$ in atoms of the form
\code{read(a,}$i$\code{)} or \code{read(b,}$i$\code{)} are aligned with
characters' stream positions:
\begin{lstlisting}[numbers=none,basicstyle=\ttfamily\footnotesize,belowskip=0mm,aboveskip=0mm]
  #iinit 0.
  #cumulative t.
  #external read(a,t+1).  #external read(b,t+1).
  accept(t) :- read(a,t), read(a,t-1), not read(a;b,t+1).
\end{lstlisting}
The (undefined) atoms appearing after the keyword ``\code{\#external}'' are declared as inputs to
\code{\#cumulative} blocks; that is, they are protected from program simplifications
until they become defined (by external rules from the controller).
Observe that any instance of the predicate \code{read}$/2$ is defined externally.
In particular, future inputs that can be provided at (schematic) step \code{t+1} are
used to cancel an obsolete rule defining \code{accept(t)},
once it does no longer refer to the last position of a stream prefix.
Given that inputs are expected from step $1={}$\code{t+1} on,
the directive ``\code{#iinit 0.}'' specifies \code{t}${}=0$ as starting value to
instantiate \code{\#cumulative} blocks for,
so that \code{read(a,1)} and \code{read(b,1)} are initially declared as external inputs.
In view of this, the answer set obtained for stream prefix $aa$ at step~$2$
includes \code{accept(2)}
(generated via 
 ``\code{accept(2)} \code{:-} 
   \code{read(a,2),}
   \code{read(a,1),}
   \code{not} \code{read(a,3),}
   \code{not} \code{read(b,3).}''),
while \code{accept(3)} does not belong to the answer set
for $aab$ at step~$3$.
Note that acceptance at different stream positions is indicated by distinct
instances of \code{accept}$/1$;
this is important to comply with modularity conditions (cf.~\cite{gegrkaobsasc12b,gegrkaobsasc12a})
presupposed by \oclingo, which essentially object to the (re)definition of (ground) head atoms
at different % incremental 
steps.

Although the presented reactive ASP encoding correctly accepts stream prefixes
matching $(a|b)^*aa$, the fact that external instances of \code{read}$/2$ are accumulated
over time is a major handicap, incurring memory pollution upon running \oclingo\ for a
(quasi) unlimited number of steps.\footnote{%
Disposal of elapsed input atoms that are yet undefined,
such as \code{read(b,1)}, \code{read(b,2)}, and \code{read(a,3)} wrt.\ stream prefix $aab$,
can be accomplished via ``\code{\#forget\ }$i$\code{.}'' directives.
Ground rules mentioning such atoms are in turn ``automatically'' simplified
by \clasp\ (cf.~\cite{gekakaosscth08a}).}
To circumvent this,
external inputs could be made transient, as in the following alternative
sequence of facts from \oclingo's controller component:
\begin{lstlisting}[numbers=none,basicstyle=\ttfamily\footnotesize,belowskip=0mm,aboveskip=0mm]
  #step 1. #volatile. read(a,1,1).
  #step 2. #volatile. read(a,1,2). read(a,2,2).
  #step 3. #volatile. read(a,2,3). read(b,3,3).
\end{lstlisting}
Note that the first two readings,
represented by \code{read(a,1,1)} and \code{read(a,2,2)}  as well as
\code{read(a,1,2)} and \code{read(a,2,3)},  are provided by facts twice,
where the respective (last) stream position is included as additional argument
to avoid redefinitions. 
In view of this,
the previous offline encoding could be replaced by the following one:
\begin{lstlisting}[numbers=none,basicstyle=\ttfamily\footnotesize,belowskip=0mm,aboveskip=0mm]
  #cumulative t.
  #external read(a,t-1;t,t).  #external read(b,t-1;t,t).
  accept(t) :- read(a,t-1,t), read(a,t,t).
\end{lstlisting}
While the automatic expiration of transient inputs after each inquiry from the controller
(along with the possible use of ``\code{\#forget\ }$i$\code{.}'' directives)
omits a blow-up in space as well as an explicit cancelation of outdated rules,
it leads to the new problem that the whole window contents (two readings in this case)
must be passed in each inquiry from the controller.
This delegates the bookkeeping of sliding window contents to the controller,
which then works around limitations of reactive ASP as introduced in~\cite{gegrkasc11a}.

Arguably, neither accumulating inputs (that are no longer inspected) over time
nor replaying window contents (the width of the window many times) is acceptable
in performing continuous ASP-based stream reasoning.
To overcome the preexisting limitations,
we introduced \emph{time-decaying logic programs}~\cite{gegrkaobsasc12b}
that allow for associating arbitrary \emph{life spans} (rather than just~$1$)
with transient program parts.
Such life spans are given by integers~\code{l} in directives of the form
``\code{#volatile : l.}'' (for online data)
or ``\code{#volatile t : l.}'' (for offline encoding parts).
With our example,
stream readings can now be conveniently passed in \code{#volatile} blocks
of life span~$2$ as follows:
\begin{lstlisting}[numbers=none,basicstyle=\ttfamily\footnotesize,belowskip=0mm,aboveskip=0mm]
  #step 1. #volatile : 2. read(a,1).
  #step 2. #volatile : 2. read(a,2).
  #step 3. #volatile : 2. read(b,3).
\end{lstlisting}
In view of automatic expiration in two steps
(eg.\ at step~$3$ for \code{read(a,1)} provided at step~$1$),
the following stripped-down offline encoding correctly handles
stream prefixes:
\begin{lstlisting}[numbers=none,basicstyle=\ttfamily\footnotesize,belowskip=0mm,aboveskip=0mm]
  #cumulative t.
  #external read(a,t).  #external read(b,t).
  accept(t) :- read(a,t-1), read(a,t).
\end{lstlisting}
Note that the embedding of encoding rules in a \code{\#cumulative} block
builds on automatic rule simplifications relative to expired inputs.
As an (equivalent) alternative, one could use ``\code{#volatile t : 2.}'' to % enforce the % an
% explicit 
% expiration of 
discard outdated rules % by means of 
via internal assumption literals
(cf.~\cite{gekakaosscth08a}).
However,
we next investigate
more adept uses of blocks for % encapsulating 
offline rules.

%% file: modeling.tex
\section{Modeling and Reasoning}
\label{sec:modeling}

The case studies provided below aim at illustrating particular features
in modeling and reasoning with time-decaying logic programs and stream data.
For the sake of clarity, we concentrate on toy domains rather than any actual
target application. % of ASP-based stream reasoning.
We begin with modelings of the simple task to monitor consecutive user accesses,
proceed with an overtaking scenario utilizing frame axioms, and then
turn to the combinatorial problem of online job scheduling;
the corresponding encodings can be downloaded at~\cite{oclingo}.
%
% The examples serve the elaboration of modeling approaches,
% rather than representing application domains.

%%% Local Variables: 
%%% mode: latex
%%% TeX-master: "paper"
%%% End: 

%% file: access.tex
\EXT{\subsection{Access Control}\label{sec:access}}{\subsection{Access Control}\label{sec:access}}

\begin{lstlisting}[float=t%
,frame=lines%
%,framesep=0pt%
,caption={Stream of user accesses with life span of~$3$ steps\newline\vspace*{-10pt}}%
,label=fig:accessStream%
,linerange={1-2,8-10,16-18,24-26,32-34,40-40,44-44,50-50,52-52,58-58,60-60}%,63-66}%
]
#step 1. #volatile : 3.
access(alice,granted,1).
#step 2. #volatile : 3.
access(alice,denied,3).
access(bob,denied,3).
#step 3. #volatile : 3.
access(alice,denied,2).
access(claude,granted,5).
#step 4. #volatile : 3.
access(bob,denied,2). access(bob,denied,4).
access(claude,denied,2).
#step 5. #volatile : 3.
access(alice,denied,4).
access(claude,denied,3). access(claude,denied,4).
#step 6. #volatile : 3.
access(alice,denied,6).
#step 7. #volatile : 3.
access(alice,denied,8).
#step 8. #volatile : 3.
access(alice,denied,7).
\end{lstlisting}
Our first scenario considers users attempting to access some service,
for instance, by logging in via a website.
Access attempts can be \code{denied} or \code{granted}, eg.\ depending on the supplied password,
and a user account is (temporarily) closed in case of three access denials in a row.
A\EXT{n (initial)}{} stream segment of access attempts is shown in Listing~\ref{fig:accessStream}.
It provides data about three users, \EXT{called }{}\code{alice}, \code{bob}, and \code{claude}.
As specified by ``\code{#volatile : 3.}'' (for each step), the life span of access data
is limited to three incremental steps
(which may be correlated to some period of real time),
aiming at an (automatic) reopening of closed user
accounts after some waiting period has elapsed.
\EXT{Furthermore, we assume}{We further assume} that time stamps\EXT{ provided}{} in the third argument of facts over 
\code{access}$/3$ deviate from\EXT{ the number}{}~$i$ in ``\code{#step\ }$i$\code{.}''
by at most~$2$; that is, the terms used in transient facts are coupled to
the step number (in an underlying incremental logic program).
Given the \EXT{stream }{}segment in Listing~\ref{fig:accessStream},
the following table summarizes non-expired logged accesses per step,
where granted accesses are enclosed in brackets and sequences of three
consecutive denials are underlined:

% \noindent
% \begin{minipage}{.6\columnwidth}
% \begin{tabular}{|r||l|l|l|}
% \cline{1-4}
% $i$ & \code{alice} & \code{bob} & \code{claude}
% \\\cline{1-4}\cline{1-4}
% $1$ & [\code{1}] & &
% \\\cline{1-4}
% $2$ & [\code{1}] \code{3} & \code{3} &
% \\\cline{1-4}
% $3$ & [\code{1}] \code{2} \code{3} & \code{3} & [\code{5}]
% \\\cline{1-4}
% $4$ & \code{2} \code{3} & \underline{\code{2} \code{3} \code{4}} & \code{2} [\code{5}]
% \\\cline{1-4}
% $5$ & \code{2} \code{4} & \code{2} \code{4} & \underline{\code{2} \code{3} \code{4}} [\code{5}]
% \\\cline{1-4}
% $6$ & \code{4} \code{6} & \code{2} \code{4} & \underline{\code{2} \code{3} \code{4}}
% \\\cline{1-4}
% $7$ & \code{4} \code{6} \code{8} & & \code{3} \code{4}
% \\\cline{1-4}
% $8$ & \underline{\code{6} \code{7} \code{8}} & &
% \\\cline{1-4}
% \end{tabular}
%
\begin{tabular}{|l||l|l|l|l|l|l|l|l|}
\hline
\multicolumn{1}{|r@{~}||}{$i$} & \multicolumn{1}{c|}{$1$} & \multicolumn{1}{c|}{$2$} & 
\multicolumn{1}{c|}{$3$} & \multicolumn{1}{c|}{$4$} & \multicolumn{1}{c|}{$5$} &
\multicolumn{1}{c|}{$6$} & \multicolumn{1}{c|}{$7$} & \multicolumn{1}{c|}{$8$}
\\\hline\hline
\code{alice} & [\code{1}] & [\code{1}] \code{3} & [\code{1}] \code{2} \code{3} & 
\code{2} \code{3} & \code{2} \code{4} & \code{4} \code{6} & \code{4} \code{6} \code{8} & 
\underline{\code{6} \code{7} \code{8}}
\\\hline
\code{bob} & & \code{3} & \code{3} & \underline{\code{2} \code{3} \code{4}} & \code{2} \code{4} & 
\code{2} \code{4} & &
\\\hline
\code{claude} & & & [\code{5}] & \code{2} [\code{5}] & \underline{\code{2} \code{3} \code{4}} [\code{5}] &
\underline{\code{2} \code{3} \code{4}} & \code{3} \code{4} &
\\\hline
\end{tabular}
% \end{minipage}%
% \begin{minipage}{.4\columnwidth}
%~\\
% \vspace{0.2\baselineskip}

\noindent
For instance, observe that
the three denied accesses by \code{bob} logged in the second and fourth
step are consecutive in view of the time stamps \code{3}, \code{2}, and~\code{4}
provided as argument values, eg.\ in
\code{access(bob,denied,3)} expiring at step~$5$.
% \end{minipage}

\begin{lstlisting}[float=t%
,frame=lines%
%,framesep=0pt%
,caption={Cumulative access control encoding\newline\vspace*{-10pt}}%
%          (\accessCumulative)}% : \lstinline{oclingo --iinit="1-offset"\ \accessCumulative}
,label=fig:accessCumulative%
,linerange={14-16,22-23,25-26,28-29,35-36,39-39,42-45,48-48,54-55,58-58}%,63-66}%
]
#const window=3. #const offset=2. #const denial=3. #iinit 1-offset.

#base.
user(bob;alice;claude). % some users
signal(denied;granted). % some signals
{ account(U,closed) : user(U) }.
account(U,open) :- user(U), not account(U,closed).

#cumulative t.
#external access(U,S,t+offset) : user(U) : signal(S).
denied(U,1,     t) :- access(U,denied,t+offset).
denied(U,N+1,   t) :- access(U,denied,t+offset),
                      denied(U,N,t-1), N < denial.
denied(U,denial,t) :- denied(U,denial,t-1).
:- denied(U,denial,t), not account(U,closed).

#volatile t.
:- account(U,closed), not denied(U,denial,t).
\end{lstlisting}%
\EXT{%
\lstinputlisting[float=t%
,frame=lines%
,aboveskip=-12.5pt%
%,framesep=0pt%
,caption={Volatile access control encoding\newline\vspace*{-10pt}}%
%          (\accessVolatile)}% : \lstinline{oclingo --iinit="2-retain"\ }\accessVolatile}%
,label=fig:accessVolatile%
,firstnumber=8%
,linerange={8-9,12-12,15-15,18-21,24-26},%linerange={14-17,23-30,36-53,59-62}%
]{examples/accessVolatile.lp}}{}%
Our first offline % incremental 
encoding 
% for determining whether a user account ought to be closed or remain open wrt.\ (non-expired) access data 
is shown in Listing~\ref{fig:accessCumulative}.
To keep the % concrete 
sliding window width, matching the life span of stream transients,
adjustable, the constant \code{window} is introduced in Line~1 (and set to default value~$3$).
Similarly, the maximum deviation of time stamps from incremental step numbers and the threshold
of consecutive denied accesses at which an account is closed are represented by the constants
\code{offset} and \code{denial}.
After introducing the % aforementioned 
three users and the possible outcomes of their access attempts
via facts, the static \code{\#base} part includes in Line~6 a choice rule on whether the status of
a user account is \code{closed}, while it remains \code{open} otherwise.
In fact, the dynamic    parts of the incremental program solve the task to identify the current
status of the account of a user~\code{U} and represent it by including either
\code{account(U,closed)} or \code{account(U,open)} in an answer set,
%
% This needs to be accomplished 
yet without redefining \code{account}$/2$ over steps.
This makes sure that step-wise (ground) incremental program parts are modularly composable
(cf.~\cite{gegrkaobsasc12b,gegrkaobsasc12a}),
which is required for meaningful closed-world reasoning by \oclingo\ in reactive settings.
% compositional,
% ie.\ that their joins are well-defined \EXT{according to }{(cf.\ }%
% Definition~\ref{def:reactive:modularity}\EXT{}{)}.
%, we must not redefine \code{account}$/2$ over steps.
% Rather, choice rules are introduced once in the static program part,
% and synchronization with data is accomplished by other means than rules
% defining \code{account}$/2$, viz.\ by posting integrity constraints.

The       encoding in Listing~\ref{fig:accessCumulative}
(mainly) relies on accumulating rules given
below ``\code{\#cumulative t.}'' (in Line~9),
% thus 
resembling incremental
planning encodings (cf.~\cite{gekakaosscth08a})
based on a  history of actions.
In order to react to external inputs, % as in \cite{gegrkasc11a},
the \code{\#external} directive in Line~10 declares
(undefined) atoms as inputs that can be provided by the environment,
ie.\ the controller component of \oclingo.
% which is a stream of user access attempts here.
Note that        instances of \code{access}$/3$ with time stamp \code{t+offset}
(where \code{offset} is the        maximum deviation from~\code{t})
are introduced    at incremental step~\code{t};
in this way, ground rules are prepared for\EXT{ (forward)}{} shifted access data
arriving early.
The rules in Line~11--13 implement the counting of consecutive denied access
attempts per user, up to the threshold given by % \EXT{ constant}{} 
\code{denial};
if this threshold is reached (wrt.\ non-expired access data),
% it means that 
the account of the respective user        is temporarily closed.
The ``positive'' conclusion  from \code{denial} many denied access attempts to
closing an account  is encoded via the integrity constraint in Line~15,
while more care is needed in concluding the opposite:
the right decision whether to leave an account open can be made only after
inspecting the whole \EXT{contents of the current window}{window contents}.
To this end, the rule in Line~14 passes information about the
threshold being reached on to later steps,
and the query in Line~18, refuting an account to be closed if there were
no three consecutive denied access attempts\EXT{ by a user}{}, is included only for
the (currently) last incremental step.

For the stream segment in Listing~\ref{fig:accessStream},
in the fourth incremental step,
we have that
\code{denied(bob,3,4)} is derived in view of (transient) facts
\code{access(bob,denied,3)}, \code{access(bob,denied,2)}, and 
\code{access(bob,denied,4)}.
% In view of 
Due to the integrity constraint in Line~15 of Listing~\ref{fig:accessCumulative},
this enforces \code{account(bob,closed)} to hold.
Since \code{access(bob,denied,3)} expires at step~$5$,
we can then no longer derive \code{denied(bob,3,4)},
and \code{denied(bob,3,5)} does not hold either;
the query in Line~18 thus enforces \code{account(bob,closed)} to be false
in the fifth incremental step.
Similarly, we have that \code{account(claude,closed)} or
\code{account(alice,closed)} hold at     step  $5$, $6$, and $8$,
respectively, but are enforced to be false at any other step.
% In fact, albeit the semantics of modules principally admits input
% atoms to assume ``arbitrary'' truth values (coherent with a   module's rules),
% \oclingo\ disambiguates this situation by fixing (undefined) input atoms 
% to be false.
% Furthermore, one may have noticed that,
As one may have noticed,
given the values of constants in Listing~\ref{fig:accessCumulative},
the consideration of atoms over \code{access}$/3$       starts at
\code{t+offset}${}=3$ for \code{t}${}=1$.
% However, rather than filling up initial contents via additional rules
% in the \code{\#base} program part, we added the ``\code{\#iinit}'' directive
% % the command-line switch \code{--iinit}
% to \oclingo\ for initializing a window wrt.\ the ``past.''
To still initialize the first (two) windows wrt.\ the ``past,''
an \code{\#iinit} directive is provided in Line~1.
By using \code{1-offset}${}=-1$ as starting value for~\code{t},
once the first external inputs are % element of an online progression is
processed, ground rules treating access data with time stamps \code{1}, \code{2}, and \code{3}
are readily available, which is exactly the range admitted at the first step.
Moreover, at a step like $6$, the range of admissible time stamps starts at~\code{4},
and it increases at later steps;
that is, inputs with time stamp~\code{3}, declared at step~$1$, are not anymore supposed to
% be provided anymore 
occur in stream data.
To enable % permanent 
program simplifications by fixing such atoms to false,
online data can be accompanied by  ``\code{\#forget\ }$i$\code{.}'' directives, % of the form 
% within its elements,
and they are utilized in practice to dispose of
elapsed input atoms (cf.~\cite{oclingo}).

\EXT{%
While, for simplicity, our first encoding accumulated rules
(simplified only in view of \code{\#forget} directives) over expiring data,
we now turn to an alternative approach in which outdated rules expire along % in parallel
with stream data.
This second encoding, shown in Listing~\ref{fig:accessVolatile},
performs the same task as the previous one,
and its first part, including constant declarations and \code{\#base} (up to Line~7), is as before.
However, all remaining rules belong to time-decaying programs of fixed life span,
making use of the observation that a time stamp~$i$ is introduced at step
\code{t}${}=i{-}$\code{offset} and last referenced at step
$i{+}$\code{offset}$+$\code{window}${-}1$.
In view of this, the (shortest) sufficient life span is
calculated in Line~8 of Listing~\ref{fig:accessVolatile} and associated
with a \code{\#volatile} program part including all rules below Line~9.
In order to avoid using another \code{\#volatile} block of life span~$1$,
as included in Listing~\ref{fig:accessCumulative},
the basic idea is to count denied access attempts downstream,
so that the oldest non-expired program part at a step has an overview
of the whole current window contents.
This approach is implemented by the rules in Line~12--15, where counting
relies on \code{denied}$/3$ instances from the next step, serving
as (additional) inputs declared in Line~11.
However, since the incremental parameter~\code{t} is included in each head atom,
we still have that the atoms defined in different steps do not overlap, so that
incremental program parts stay compositional.
To make the right decisions about whether to close accounts in view of the
outcomes of consecutive denial counting,
the integrity constraint in Line~16 can be used unchanged
(cf.\ Line~15 in Listing~\ref{fig:accessCumulative}).
Unlike this,
the integrity constraint in Line~17, % --18,
enforcing atoms of the form \code{account(U,closed)} to be false,
now checks renouncement of the \code{denial} threshold over all (possibly) non-expired 
preceding program parts;
in this way, the oldest program part is guaranteed to be
among the investigated ones.%
% Finally, note that the encoding in Listing~\ref{fig:accessVolatile}
% relies on a filled window to avoid free choices over % instances of 
% \code{account}$/2$;
% initially, this is achieved via ``\code{#iinit 2-retain.}''
}{%
% can be achieved via \oclingo\ switch \code{--iinit}${}={}$``$2{-}$\code{retain}.''}{%
}

\EXT{%
Both encodings presented before have the drawback that the knowledge-intensive logic program part
% modeling the task at hand 
                          is (gradually) replaced at each step.}{%
In addition to the cumulative encoding in Listing~\ref{fig:accessCumulative},
we also devised a volatile variant
(cf.\ extended draft~\cite{gegrkaobsasc12a})
in which outdated rules expire along with stream data.
Both the cumulative and the volatile encoding have the drawback that the knowledge-intensive logic program part
% modeling a task at hand 
                        is (gradually) replaced at each step.}
While this is tolerable in the simple access scenario, 
for more complex problems, it means 
that the internal problem representation of a solving component like \clasp\ changes
significantly upon processing stream data, 
so that a potential re-use of learned constraints is limited.
In order to reinforce conflict-driven learning,
we next demonstrate a modeling approach
allowing for the preservation of a static problem representation in
view of fixed window capacity.

At the beginning (up to Line~5),
our static access control encoding, shown in Listing~\ref{fig:accessBase},
is similar to the cumulative approach (Listing~\ref{fig:accessCumulative}),
while the \code{\#base} part is significantly extended in the sequel.
In fact, the constant \code{modulo}, calculated in Line~6, takes into account
that at most \code{window} many consecutive online inputs % elements of an online progression
are jointly available (preceding ones are expired) and that terms representing
time stamps may deviate by up to \code{offset} (both positively and negatively)
from incremental step numbers.
Given this, % note that
\code{window}$+2*$\code{offset} slots are sufficient to accommodate all distinct
time stamps provided as argument values in instances of \code{access}$/3$
belonging to a window,
and one additional slot is added in Line~6 as   separator between the largest
and the smallest (possibly) referenced time stamp.
The available slots are then arranged in a cycle (via modulo arithmetic) in Line~7,
and the counting of denied access attempts per user, implemented
in Line~9--11, traverses consecutive slots according to instances of the predicate \code{next}$/2$.
Importantly,     counting does not rely on transient external inputs,
ie.\ % instances of 
\code{access}$/3$, but instead refers to instances of \code{baseaccess}$/3$,
provided by the choice rule in Line~8;
in this way, the atoms and rules in the \code{\#base} program part can be re-used in
determining the status of user accounts wrt.\ transient access data from a stream.

\begin{lstlisting}[float=t%
,frame=lines%
%,framesep=0pt%
,caption={Static access control encoding\newline\vspace*{-10pt}}%
%          (\accessBase)}% : \lstinline{oclingo --iinit="2-modulo"\ }\accessBase}%
,label=fig:accessBase%
,firstnumber=6%
,linerange={18-18,31-31,34-34,39-41,43-44,50-51,54-54,57-58,64-65,69-70}%,linerange={14-17,23-39,45-52,58-62,69-72}%
]
#const modulo=window+2*offset+1. #iinit 2-modulo.
next(T,(T+1) #mod modulo) :- T := 0..modulo-1.
{ baseaccess(U,denied,T) : user(U) : next(T,_) }.
denied(U,1,  T) :- baseaccess(U,denied,T).
denied(U,N+1,S) :- baseaccess(U,denied,S), next(T,S),
                   denied(U,N,T), N < denial.
account(U,closed) :- denied(U,denial,T).
account(U,open)   :- user(U), not account(U,closed).

#cumulative t.
#external access(U,S,t+offset) : user(U) : signal(S).
:- access(U,denied,T), T := t+offset,
   not baseaccess(U,denied,T #mod modulo).

#volatile t : modulo.
:- baseaccess(U,denied,(T+modulo) #mod modulo),
   not access(U,denied,T), T := t+offset.
#const window=3.                 % external events in window for 3 steps (by default)
#const offset=2.                 % time points/terms may deviate up to 2 (by default)
% #const modulo=window+2*offset+1. % number of time points in base program (calculated)
#const denial=3.                 % account closed when 3 accesses denied (by default)
% BASE %
%%%%%%%%

#base.

user(bob;alice;claude). % some users
signal(denied;granted). % some signals

next(T,(T+1) #mod modulo) :- T := 0..modulo-1.
% successors in fixed-size time circle

{ baseaccess(U,denied,T) : user(U) : next(T,_) }.
% guess base window contents
% :- baseaccess(U,S,T), S != denied.                       % use "denied" accesses only

% count consecutive denied user accesses along time circle up to a "denial" threshold
denied(U,1,  T) :- baseaccess(U,denied,T).
             % account closed if at threshold  % or open otherwise

%%%%%%%%%%%%%%
% CUMULATIVE %
%%%%%%%%%%%%%%

#cumulative t.

   not baseaccess(U,denied,T #mod modulo).

%%%%%%%%%%%%
% VOLATILE %
:- baseaccess(U,denied,(T+modulo) #mod modulo),
   not access(U,denied,T), T := t+offset.
% "modulo" added above to get non-negative number when #mod applied w.r.t. past event
\end{lstlisting}
To get decisions on closing accounts right
(via the rules in Line~12--13),
synchronization between instances of \code{baseaccess}$/3$ and (transient) facts over
\code{access}$/3$ is implemented in the \code{\#cumulative} and \code{\#volatile}
parts in Listing~\ref{fig:accessBase}.
Beyond declaring inputs in the same way as before
(cf.\ Line~10 in Listing~\ref{fig:accessCumulative}),
for any non-expired online input of the form \code{access(U,denied,t+offset)}, % given in an online progression,
the integrity constraint in Line~17--18 enforces the corresponding instance
of \code{baseaccess}$/3$, % determined by calculating
calculated via
``\code{(t+offset) \#mod modulo},'' to hold.
Note that this constraint does not need to be expired explicitly
(yet it could      be moved to the \code{\#volatile} block below)
because elapsed input atoms render it 
% instances 
ineffective anyway.
However, % instances of 
the integrity constraint in Line~21--22,
enforcing \code{baseaccess}$/3$ counterparts of non-provided facts of the form
\code{access(U,denied,t+offset)} to be false,
must be discharged once the sliding window progresses
(by \code{modulo} many steps).
For instance, when \code{offset}${}=2$ and \code{modulo}${}=8$,
input atoms of the form \code{access(U,denied,3)},
introduced at the first step, map to
\code{baseaccess(U,denied,3)}, and the same applies to
instances of \code{access(U,denied,11)},
becoming 
available at the ninth incremental step.
Since the smallest time stamp that can be mentioned by
non-expired inputs at the ninth step is~\code{5}
(inputs given before step~$7$ are expired),
the transition of % ground 
integrity constraints
(mapping time stamp \code{11} instead of \code{3} to slot \code{3})
is transparent.
In addition,
as % unavailable 
expired inputs with time stamp~\code{4} enforce
atoms of the form \code{baseaccess(U,denied,4)}
to be false (via % instances of 
the integrity constraint in Line~21--22 instantiated for step~$2$),
denial counting in Line~9--11 stops at
     atoms \code{baseaccess(U,denied,3)},
representing latest stream data at the ninth step.
Hence, \code{denial} many consecutive instances of \code{baseaccess(U,denied,T)},
needed to close the account of a user~\code{U}, correspond to
respective 
facts over \code{access}$/3$ in the current window.
Finally, to
avoid initial guesses over \code{baseaccess}$/3$ wrt.\ (non-existing)
denied accesses lying in the past,
``\code{#iinit 2-modulo.}'' is included in Line~6.
% \oclingo\ ought to be launched with % the switch 
% \code{--iinit}${}={}$``$2{-}$\code{modulo}.''
Then, 
instances of \code{baseaccess}$/3$
for (positive) values
``\code{(T+modulo) #mod modulo},'' calculated 
in Line~21,
% temporarily enforcing 
are (temporarily) enforced
 to be false
when they match
                   \code{access}$/3$ instances
with non-positive time stamps.

%%% Local Variables: 
%%% mode: latex
%%% TeX-master: "paper"
%%% End: 

%% file: traffic.tex
\EXT{\subsection{Overtaking Maneuver Recognition}\label{sec:traffic}}{\subsection{Overtaking Maneuver Recognition}\label{sec:traffic}}

\begin{figure}[t]
  \caption{Automaton for recognizing overtaking maneuvers%;
%           completion indicated by output ``OT!'' at state $F$
           \label{fig:atm_ot}}
%Illustration of the course of events for a successful
%    overtake maneuver indicated by output ``OT!'' at state f.}
  \centering
    \resizebox{0.7\linewidth}{!}{%
      \begin{tikzpicture}[->,>=stealth',shorten >=1pt,auto, node
        distance=3cm,semithick]
        \tikzstyle{every state}=[draw=black,text=black]
        \node[state,initial]             (I)                    {$\varnothing$};
        \node[state]                     (b) [right of=I] {$B$};
        \node[state]                     (n) [right of=b] {$N$};
        \node[state with output]         (f) [below of=b] {$F$\nodepart{lower} ``OT!''};

        \path (I) edge [bend left]         node {behind}             (b)
            edge [loop above]        node {infront,nextto,$\varepsilon$}()
        (b) edge [bend left]         node {nextto}                   (n)
            edge [loop above]        node {behind,$\varepsilon$}        ()
            edge [bend left]         node {infront}                  (I)
        (n) edge                     node {infront}                  (f)
            edge [loop right]        node {nextto,$\varepsilon$}        ()
            edge [bend  left]        node {behind}                   (b)
        (f) edge                     node {infront,nextto,$\varepsilon$}(I)
            edge                     node {behind}                   (b);
      \end{tikzpicture}
    }%
%  \end{centering}
% \vertgap
\end{figure}
Our second scenario deals with recognizing the completion of overtaking maneuvers
by a car, eg.\ for signaling it to the driver.
The recognition follows the transitions of the automaton in Figure~\ref{fig:atm_ot}.%
\footnote{%
Unlike in this simple example scenario,
transition systems are usually described compactly 
in terms of state variables and operators on them,
eg.\ defined via action languages (cf.~\cite{gilelimctu03a}).
The automaton induced by a compact description can be of exponential size,
and an explicit representation like in Figure~\ref{fig:atm_ot} is
often inaccessible in practice.}
Starting from state~$\varnothing$, representing that a maneuver has not yet been initiated,
sensor information about being ``behind,'' ``nextto,'' or ``infront'' of another car
enables transitions to corresponding states~$B$, $N$, and~$F$.
As indicated by the output ``OT!'' in~$F$,
an overtaking maneuver is completed when~$F$ is reached from~$\varnothing$
via a sequence of ``behind,'' ``nextto,'' and ``infront'' signals.
Additional $\varepsilon$ transitions model the progression from one time point to the next
in the absence of signals:
while such transitions are neutral in     states~$\varnothing$, $B$, and~$N$,
the final state~$F$ is abandoned (after outputting ``OT!'').
For instance,
the automaton in Figure~\ref{fig:atm_ot}
admits the following trajectory
(indicating a state at time point~$i$ by ``$@i$''
 and providing signals in-between states):
\pagebreak[1]%
\begin{align*}
\lackgap
(&\varnothing@0,\linebreak[1]
 \textnormal{behind},\linebreak[1]B@1,\linebreak[1]
 \varepsilon,\linebreak[1]B@2,\linebreak[1]
 \textnormal{nextto},\linebreak[1]N@3,\linebreak[1]
 \textnormal{infront},\linebreak[1]F@4,\linebreak[1]
 \varepsilon,\linebreak[1]
{}\breakeq 
% \varepsilon,\linebreak[1]{}& \varnothing@4,\linebreak[1]
\lackgap
&
% \textnormal{infront},
 \varnothing@5,\linebreak[1]
 \textnormal{nextto},\linebreak[1]\varnothing@6,\linebreak[1]
 \textnormal{behind},\linebreak[1]B@7,\linebreak[1]
 \textnormal{nextto},\linebreak[1]N@8,\linebreak[1]
 \varepsilon,\linebreak[1]N@9).
\end{align*}
Here, an overtaking maneuver is completed when~$F$ is reached at
time~$4$, and $\varepsilon$ transitions preserve~$B$ and~$N$, but not~$F$.
In the following,
we consider overtaking maneuvers that are completed in at most~$6$ steps;
that is, for a given time point~$i$, the automaton in Figure~\ref{fig:atm_ot}
is assumed to start from~$\varnothing$ at time $i{-}6$.

Similar to the static access control encoding in Listing~\ref{fig:accessBase},
our encoding of overtaking maneuver recognition,
shown in Listing \ref{lst:overtakingBase},
uses modulo arithmetic to map time stamps in stream data to a corresponding
slot of atoms and rules provided in the \code{\#base} program part.
In more detail, transient (external) facts of the form
\code{at(P,C,T)} (in which \code{P} is 
\code{behind}, \code{nextto}, or \code{infront} and
\code{C} refers to a
\code{red}, \code{blue}, or \code{green} car) are matched with
corresponding instances of \code{baseat}$/3$ by means of the
\code{\#cumulative} and \code{\#volatile} parts in Line~20--24.
As before, these parts implement a transparent shift from
steps~$i$ to~$i+$\code{modulo},
provided that transient stream data is given in % partitioned into
``\code{\#volatile : modulo.}'' blocks.
Then, the rules in Line~12--16 model state transitions based on signals,
and the one in Line~17--18 implements $\varepsilon$ transitions
(making use of projection in Line~10)
as specified by the automaton in Figure~\ref{fig:atm_ot}.
In particular, the completion of an overtaking maneuver in the
current step, indicated by deriving \code{infront} as state
for a car~\code{C} and a time slot~\code{S} via the rule in Line~15--16,
relies on \code{now(S)} (explained below).
Also note that state~$\varnothing$ is left implicit, ie.\
it applies to a car~\code{C} and a slot~\code{T}
if \code{baseat(P,C,T)} does not hold for any relative position~\code{P},
and that the frame axioms represented in Line~17--18
do not apply to \code{infront} states.

\begin{lstlisting}[float=t%
,frame=lines%
%,framesep=0pt%
,caption={Static overtaking maneuver recognition encoding\newline\vspace*{-10pt}}%(\overtakingBase}% : \lstinline{oclingo --iinit="2-modulo"\ }\jobscheduleB}%
,label=lst:overtakingBase%
%,linerange={}
]
#const modulo=6. #iinit 2-modulo.
#base.
position(behind;nextto;infront). % relative positions
car(red;blue;green).             % some cars
time(0..modulo-1).               % time slots

next(T,(T+1) #mod modulo) :- time(T), not now(T).
{ now(T) : time(T) } 1.
{ baseat(P,C,T) : position(P) : car(C) : time(T) }.
baseat(C,T) :- baseat(_,C,T).

state(behind, C,T) :- baseat(behind,C,T).
state(nextto, C,S) :- baseat(nextto,C,S), next(T,S),
                      1 { state(behind,C,T), state(nextto,C,T) }.
state(infront,C,S) :- baseat(infront,C,S), now(S),
                      next(T,S), state(nextto,C,T).
state(P,      C,S) :- state(P,C,T), P != infront,
                      next(T,S), not baseat(C,S).

#cumulative t.
#external at(P,C,t) : position(P) : car(C).
:- at(P,C,t), not baseat(P,C,t #mod modulo).
#volatile t : modulo.
:- baseat(P,C,(t+modulo) #mod modulo), not at(P,C,t).
#volatile t.
:- not now(t #mod modulo).
\end{lstlisting}
While a \code{next}$/2$ predicate had also been defined in Listing~\ref{fig:accessBase}
to arrange the time slots of the \code{\#base} program in a cycle,
the corresponding rule in Line~7 of Listing~\ref{lst:overtakingBase}
relies on the absence of \code{now(T)} for linking a time
slot~\code{T} to ``\code{(T+1) \#mod modulo}.''
In fact,
instances of \code{now}$/1$ are provided by the choice rule in Line~8
and synchronized with the incremental % parameter
                                   step counter~\code{t} via the integrity constraint
``\code{:- not now(t #mod modulo).}'' of life span~$1$ (cf.\ Line~25--26).
% (By using integrity constraints rather than rules for continuous synchronization,
%  we guarantee the compositionality of incremental program parts.)
Unlike the previous approach to access counting (introducing an empty slot),
making the current time slot explicit enables the linearization of a time slot cycle
also in the presence of frame axioms,
which could propagate into the ``past'' otherwise.
In fact, if prerequisites regarding \code{now}$/1$ were dropped in Line~7 and~15,
one could, beginning with % the input 
\code{at(behind,C,7)} as input and its corresponding atom
\code{baseat(behind,C,1)},
derive \code{state(infront,C,4)} at step~$7$
for a car~\code{C} subject to the trajectory given    above.
Such a conclusion is clearly unintended,
and the technique in Line~7--8 and 25--26,
using \code{now}$/1$ to linearize a time slot cycle,
provides a general solution for this issue.
Finally, % note that
% \oclingo\ ought to be launched with % the switch 
% \code{--iinit}${}={}$``$2{-}$\code{modulo}''
``\code{#iinit 2-modulo.}''
is again included  in Line~1  to avoid initial guesses
% to also avoid guesses 
over \code{baseat}$/3$ wrt.\ (non-existing)
past signals.

%% file: onlinejobs.tex
\EXT{\subsection{Online Job Scheduling}\label{subsec:onlinejobs}}{\subsection{Online Job Scheduling}\label{subsec:onlinejobs}}

After inspecting straightforward % stream 
data evaluation tasks,
we now turn to a combinatorial problem in % the problem of job scheduling,
% where 
which job requests of different durations must be scheduled
to machines without overlapping one another.
Unlike in offline job scheduling~\cite{brucker07},
where % all 
requests are known in advance,
we here assume a stream of job requests,
provided via (transient) facts % of the form
\code{job(I,M,D,T)} such that  % in which
\code{I} is a job ID,
\code{M} is a machine,
\code{D} is a duration, and
\code{T} is the arrival time of a   request.
In addition, we assume a deadline \code{T+max\_step},
for some integer constant \code{max\_step},
by which the execution of a job~\code{I} submitted at step~\code{T}
must be completed.
For instance, an (initial) segment of a job request stream can be as follows:
\begin{lstlisting}[numbers=none,basicstyle=\ttfamily\footnotesize,belowskip=0mm,aboveskip=0mm]
  #step 1 : 0.   #volatile : 21.               job(1,1,1,1).
  job(2,1,5,1).  job(3,1,5,1).  job(4,1,5,1).  job(5,1,5,1).
  #step 21 : 0.  #volatile : 21.
  job(1,1,5,21). job(2,1,5,21). job(3,1,5,21). job(4,1,5,21).
\end{lstlisting}
%
% \begin{equation*}%\label{eq:ex:abba}
% \begin{array}{@{}l@{}l@{}l@{}l}
% %\multicolumn{2}{@{}l}{%
% \text{\fcodeeq{#step 1 : 0.\ }} &
% \text{\rlap{\fcodeeq{#volatile : 21.}}}%}
% \breakeq
% \text{\fcodeeq{job(1,1,1,1).}} &
% \text{\fcodeeq{job(2,1,5,1).}} &
% \text{\fcodeeq{job(3,1,5,1).}} &
% \text{\fcodeeq{job(4,1,5,1).}} 
% % \breakeq % &
% \text{\fcodeeq{job(5,1,5,1).}}
% \\
% %%\multicolumn{2}{@{}l}{%
% \text{\fcodeeq{#step 21 : 0.}} &
% \text{\rlap{\fcodeeq{#volatile : 21.}}}%}
% \breakeq
% \text{\fcodeeq{job(1,1,5,21).}} &
% \text{\fcodeeq{job(2,1,5,21).}} &
% % \breakeq
% \text{\fcodeeq{job(3,1,5,21).}} &
% \text{\fcodeeq{job(4,1,5,21).}}
% \end{array}
% \end{equation*}
%
That is, five jobs with ID~\code{1} to~\code{5}
(of durations~\code{1} and~\code{5}) are submitted at step~$1$
and ought to be completed on machine~\code{1} within the deadline
$1{+}$\code{max\_step}${}=21$ (taking   \code{max\_step}${}=20$).
Four more jobs with ID~\code{1} to~\code{4} of duration~\code{5},
submitted at step~$21$,
also need to be executed on machine~\code{1}.
As a matter of fact, a schedule to finish all jobs within their deadlines
must first launch the five jobs submitted at step~$1$,
thus occupying machine~\code{1} at time points up to~$21$,
before the other jobs can use machine~\code{1} from time point~$22$ to~$41$.
However,
when a time-decaying logic program 
does not admit any answer set at some step
(ie.\ if there is no schedule meeting all deadlines),
% at a stream position is unsatisfiable
% (in this context it means there is no possible schedule meeting deadlines),
the default behavior of \oclingo\ is to increase the incremental 
step counter  % parameter
until an answer set is obtained.
% found instead of outputting an unsatisfiable result.
% This behavior may lead to undesired results in stream reasoning applications
% since increasing the parameter shifts the current window. 
This behavior would lead to the expiration of pending job requests,
so that a schedule generated in turn lacks part of the submitted jobs.
% We can alter this behavior by explicitly specifying the number of times, 
% which \oclingo\ is allowed to increment the parameter
% in case of an unsatisfiable result.
Since such (partial) schedules are unintended here,
we take advantage of the enriched directive ``\code{#step} $i$ \code{:} $\delta$.''
to express that increases of the step counter must not exceed $i{+}\delta$,
regardless of whether an answer set has been obtained at step $i{+}\delta$
(or some greater step).
In fact, 
since $\delta=0$ is used above,
\oclingo\ does not increase the step counter beyond~$i$, but rather
returns ``unsatisfiable'' as result if there is no answer set.

\begin{lstlisting}[float=t%
,frame=lines%
%,framesep=0pt%
,caption={Static online job scheduling encoding\newline\vspace*{-10pt}}%
%          (\jobscheduleBase)}% : \lstinline{oclingo --iinit="2-modulo"\ }\jobscheduleB}%
,label=fig:jobscheduleBase%
,linerange={6-6,8-8,11-11,13-13,17-17,19-20,25-25,28-31,33-33,71-73,76-76,79-80}
]
#const max_duration = 5. #const max_jobid = 5. #const num_machines = 5.
#const max_step = 20. #const modulo = 2*max_step+1. #iinit 2-modulo.
#base.
duration(1..max_duration). jobid(1..max_jobid). machine(1..num_machines).
next(T,(T+1) #mod modulo) :- T := 0..modulo-1.

{ basejob(I,M,D,T) : jobid(I) : machine(M) : duration(D) : next(T,_) }.
1 { jobstart(I,T,(T..T+max_step+1-D) #mod modulo) } 1 :- basejob(I,_,D,T).

occupy(M,I,T,D,  S) :- basejob(I,M,D,T), jobstart(I,T,S).
occupy(M,I,T,D-1,S) :- occupy(M,I,T,D,R), next(R,S), D > 1.
occupy(M,I,T,S)     :- occupy(M,I,T,_,S).
:- occupy(M,I1,T1,S), occupy(M,I2,T2,S), (I1,T1) < (I2,T2).

#cumulative t.
#external job(I,M,D,t) : jobid(I) : machine(M) : duration(D).
:- job(I,M,D,t), not basejob(I,M,D,t #mod modulo).
#volatile t : modulo.
:- basejob(I,M,D,(t+modulo) #mod modulo), not job(I,M,D,t).
\end{lstlisting}
In Line~1--4,
our (static) job scheduling encoding, shown in Listing~\ref{fig:jobscheduleBase},
defines the viable durations, IDs of jobs requested per step, and
the available machines in terms of corresponding constants.
Furthermore, deadlines for the completion of jobs are 
obtained by adding \code{max\_step} (set to~$20$ in Line~2)
to request submission times.
As a consequence, jobs with submission times~$i\leq j$ such that
$j\leq i{+}$\code{max\_step} may need to be scheduled jointly,
and the minimum window width \code{modulo} required to accommodate
the (maximum) completion times of jointly submitted jobs
is calculated accordingly in Line~2.
% In fact, 
Given the value of \code{modulo},
         the time slots of the \code{\#base} program part
are in Line~5 again arranged in a cycle (similar to access counting in Listing~\ref{fig:accessBase}).
%,
%and predicate \code{hzon}$/2$, defined in Line~8, gives the 
%horizon of a time slot~\code{T}, 
%ie.\ the time slots
%at which a job submitted at% time~
%~\code{T} can be scheduled for execution.
The technique applied in Line~15--19
to map job requests given by transient (external) facts over
\code{job}$/4$ to corresponding instances of \code{basejob}$/4$,
provided by the choice rule in Line~7,
remains the same as in previous static encodings
(cf.\ Listing~\ref{fig:accessBase} and~\ref{lst:overtakingBase}).
% However, 
But      note that job IDs can be shared between jobs submitted at different steps,
so that pairs \code{(I,T)} of an ID~\code{I} and a % time 
slot~\code{T}
identify  job requests uniquely in the sequel.

% Additionally, we use a maximal timespan for a schedule
% (declared as \code{max\_steps} in Line~3). 
% %This may cause that there might be some unscheduled jobs in a schedule.
% Similar to the static encoding scheme described in Section~\ref{sec:access},
% the choice rule for the \code{basejob} predicate (Lines~11-12)
% in the static part
% captures all possible job requests which can arrive within a frame of
% \code{modulo} time points. 
% At any time point we consider \code{max\_steps} steps in past to take 
% the previously arrived jobs into account and \code{max\_steps} steps
% in future to find a schedule.
% That is why, 
% \code{modulo} is calculated as $2 \times max\_steps + 1$ 
% time points in Line~3.\footnote{Consider that $n+1$ time points are needed
% to represent $n$ steps.}
% The static encoding scheme ensures that the solver sees the same general problem and
% external inputs are mapped to corresponding slots in the 
% general problem throughout the stream.
% Particullarly, the constraints in Lines~33-40 synchronize actual job requests 
% in the stream (\code{job} instances) with 
% corresponding job requests in the static part (\code{basejob} instances).

The rules in Line~8--13 of the \code{\#base} program
accomplish the non-overlapping scheduling of submitted jobs
such that they are completed within their deadlines.
In fact, the choice rule in Line~8 expresses that a job
of duration~\code{D} with submission time slot~\code{T} 
must be launched such that its execution finishes
at slot ``\code{(T+max\_step) #mod modulo}'' (at the latest).
Given the slots at which jobs are started,
the rules in Line~10--12 propagate the occupation of machines~\code{M}
wrt.\ durations~\code{D},
and the integrity constraint in Line~13 makes sure that
the same machine is not occupied by distinct jobs at the same time slot.
For instance, the following atoms of an answer set represent
starting times such that the jobs requested in the % (initial) 
stream segment
given above do not overlap and are processed within their deadlines:
\begin{lstlisting}[numbers=none,basicstyle=\ttfamily\footnotesize,belowskip=0mm,aboveskip=0mm]
  jobstart(1,1,1) 
  jobstart(2,1,2)   jobstart(1,21,22)
  jobstart(3,1,7)   jobstart(2,21,27)
  jobstart(4,1,12)  jobstart(3,21,32)
  jobstart(5,1,17)  jobstart(4,21,37)
\end{lstlisting}
% \begin{equation*}%\label{eq:ex:abba}
% \begin{array}{@{}l@{}l@{}l@{}l}
% \text{\fcodeeq{jobstart(1,1,1)\ }} &
% \text{\fcodeeq{jobstart(2,1,2)}} & 
% % \breakeq
% \text{\fcodeeq{jobstart(3,1,7)\ }} &
% %\breakeq
% \text{\fcodeeq{jobstart(4,1,12)}}
% \breakeq
% \text{\fcodeeq{jobstart(5,1,17)}}
% \\
% \text{\fcodeeq{jobstart(1,21,22)}} &
% \text{\fcodeeq{jobstart(2,21,27)}} &
% \text{\fcodeeq{jobstart(3,21,32)}} &
% %\breakeq
% \text{\fcodeeq{jobstart(4,21,37)}}
% \end{array}
% \end{equation*}
%
Note that the execution of the five jobs submitted at step~$1$
is finished at     their common deadline~$21$,
and the same applies wrt.\ the deadline~$41$ of jobs submitted at
step~$21$.
Since machine~\code{1} is occupied at all time slots,
executing all jobs within their deadlines would no longer be feasible
if a         job request like    \code{job(5,1,1,21)} were added.
In such a case, 
\oclingo\     outputs ``unsatisfiable'' and % no answer set and 
waits for    new  online input,
which may shift the window and relax the next query
% for scheduling with 
due to the expiration of some job requests.
%
% Leaving a submitted job request unprocessed is in principle
% admitted by the encoding in Listing~\ref{fig:jobscheduleBase}.
% However, the integrity constraints in Line~25--28 (along with the rule in Line~24)
% limit the dissipation of available resources by stipulating that a job is executed
% if it can finish within its deadline (Line~25--26) and that it should be launched
% at the first time slot at which the required machine is unoccupied (Line~27--28).
% Although such requirements already lead to reasonable schedules,
% harder constraints would need to be posted for insisting on the execution of all jobs.
% Then, queries for schedules may turn out to be unsatisfiable;
% implementing the functionality required for handling such cases in
% \oclingo\ is ongoing work.

%A second
A future extension of \oclingo\ regards % the support of
optimization
(via \code{\#minimize}/\code{\#maximize}) wrt.\ online data,
given that % optimal 
solutions violating as few (soft) constraints as possible
may be more helpful than just 
reporting unsatisfiability.
In either mode of operation,
the static representation of problems over a window of fixed width,
illustrated in Listing~\ref{fig:accessBase}, \ref{lst:overtakingBase},
and~\ref{fig:jobscheduleBase},
enables the re-use of constraints learned upon solving a query for
answering further queries asked later on.
% Also note that constraints involving transient atoms are simplified
% (by \oclingo's solving component) once the atoms expire,
% which can be exploited in using cumulative rules not subject to ``assumptions''
% (cf.\ \cite{gekakaosscth08a}) instead of volatile ones.
% Investigating whether this can boost the performance of stream reasoning
% is an interesting subject to future work.
% However, preliminary stress tests on large amounts of stream data also showed that the names of
% expired (ie.\ permanently false) atoms need to be disposed of at the implementation level,
% while \oclingo\ currently still remembers these names. % of atoms.
% Such tests were first enabled by the provision of modeling capacities for
% stream reasoning, thus outlining issues for future improvements.

\begin{figure}[t]%
\caption{Experimental results for online job scheduling}%
\subfigsqueeze%
\subfigure[Comparison of encoding variants on data \rlap{streams}]{\label{tab:tests}\begin{minipage}{0.494\textwidth}\input{tab-tests}\end{minipage}} \ 
\subfigure[Time plot for
``5x3x5\_15\_1'']{\label{fig:plotJobS}\includegraphics[width=0.5\textwidth]{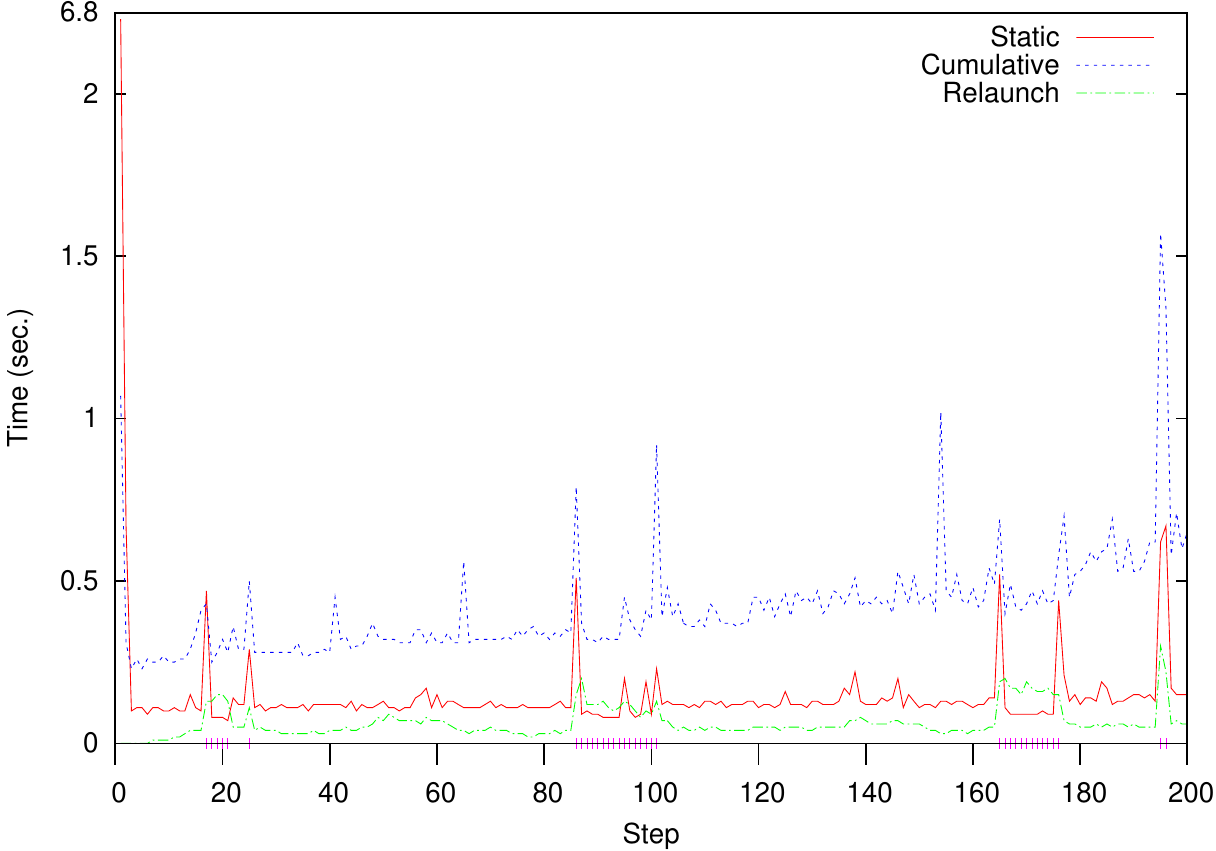}}
\subfigsqueeze%
\label{fig:tests}%
\end{figure}
Although \oclingo\ is still in a prototypical state,
we performed some preliminary experiments in order to give
an indication of the impact of different encoding variants.
As the first two example scenarios model pure data evaluation tasks
(not requiring search), 
experiments with them did not exhibit significant runtimes,
and we thus focus on results for online job scheduling.
In particular,
we assess  \oclingo\ on the static encoding in Listing~\ref{fig:jobscheduleBase}
as well as a cumulative variant
(analog to the cumulative access control encoding in Listing~\ref{fig:accessCumulative});
we further consider the standard ASP system \clingo, processing each query independently
via relaunching wrt.\ the current window contents.
Table~\ref{tab:tests}
provides average runtimes of the investigated configurations in seconds,
grouped by satisfiable ($\varnothing$S) and unsatisfiable ($\varnothing$U) queries,
on  12 randomly generated data streams with 200 % elements 
online inputs each.
These streams vary in the values used for constants,
eg.\ \code{max_jobid}${}=5$, \code{max_duration}${}=3$, \code{num_machines}${}= 5$,
and \code{max_step}${}=15$ with the three ``5x3x5\_15\_$n$'' streams,
and the respective numbers of satisfiable (\#S) and unsatisfiable (\#U) queries.
First of all,
we observe that the current prototype version of \oclingo\ cannot yet compete with
\clingo.
The reason for this is that \oclingo's underlying grounding and solving components
were not designed with expiration in mind, so that they currently still remember
the names of expired atoms (while irrelevant constraints referring to them are truly deleted).
The resulting low-level overhead in each step explains the advantage of
relaunching \clingo\ from scratch.
When comparing \oclingo's performance  wrt.\ encoding variants,
the static encoding appears to be generally more effective than
its cumulative counterpart,
albeit some unsatisfiable queries
stemming from the last three example streams in Table~\ref{tab:tests}
are solved faster using the latter.

         The plot in Figure~\ref{fig:plotJobS} provides a more fine-grained
picture by displaying runtimes for individual queries from stream ``5x3x5\_15\_1,''
where small bars on the x-axis indicate unsatisfiable queries.
While the static encoding yields a greater setup time of  \oclingo\ at the
very beginning, it afterwards dominates the cumulative encoding variant,
which requires the instantiation and integration of rules unrolling the horizons of
new job requests at each step.
Unlike this, the static encoding merely maps input atoms to their
representations in the \code{#base} part,
thus also solving each query wrt.\ the same (static) set of atoms.
As a consequence, after initial unsatisfiable queries
(yielding spikes in all configurations' runtimes),
\oclingo\ with the static encoding is sometimes able to outperform \clingo\
for successive queries remaining unsatisfiable.
In fact, when the initial reasons for unsatisfiability remain in the window,
follow-up queries are rather easy given the previously learned constraints,
and we observed that some of these queries could actually be solved without any guessing.

%% file: tab-tests.tex
%\begin{table}[t]
%   \centering
% {\normalsize 
%  \caption{Benchmark results for online job scheduling}
\scriptsize
  \begin{tabular}{@{}|l|@{\,}r@{\,}r||@{\,}r@{\,}r|@{\,}r@{\,}r|@{\,}r@{\,}r|@{}}
    \hline
     &  &  &
\multicolumn{2}{@{}c|@{\,}}{Static}&\multicolumn{2}{@{}c|@{\,}}{Cumulative}&\multicolumn{2}{@{}c|@{}}{Relaunch}\\
%    \raisebox{1.5ex}[0cm][0cm]{Stream} & \raisebox{1.5ex}[0cm][0cm]{\#s} &
% \raisebox{1.5ex}[0cm][0cm]{\#u} 
    Stream & \#S & \#U & $\varnothing$S & $\varnothing$U & $\varnothing$S & $\varnothing$U & $\varnothing$S & $\varnothing$U\\
    \hline%\multicolumn{9}{@{}c@{}}{}\\[-6.5pt]
    \hline
% \multicolumn{2}{c|}{Static}&\multicolumn{2}{c|}{Cumulative}&\multicolumn{2}{c|}{Re-launching}\\
%     \raisebox{1.5ex}[0cm][0cm]{Stream} & \raisebox{1.5ex}[0cm][0cm]{\#s} &
% \raisebox{1.5ex}[0cm][0cm]{\#u} & sat & unsat & sat & unsat & sat & unsat\\
%     \hline
%     \hline
    \input{data-tests}
    \hline
  \end{tabular}%
%   \caption{Benchmark results for the online job scheduling problem. 
% Time values are average run times.}
%   \label{tab:tests}
% }
%\end{table}%
%%% Local Variables: 
%%% mode: latex
%%% TeX-master: "paper"
%%% End: 

%% file: data-tests.tex
3x3x3\_9\_1   & 162 & 38 & 0.02 & 0.01   & 0.05 & 0.05   & 0.01 & 0.01  \\
3x3x3\_9\_2   & 151 & 49 & 0.02 & 0.01   & 0.05 & 0.04   & 0.01 & 0.01  \\
3x3x3\_9\_3   & 147 & 53 & 0.02 & 0.01   & 0.05 & 0.05   & 0.01 & 0.01  \\
5x3x5\_15\_1  & 164 & 36 & 0.17 & 0.18   & 0.40 & 0.48   & 0.05 & 0.15  \\
5x3x5\_15\_2  & 167 & 33 & 0.17 & 0.47   & 0.40 & 0.64   & 0.04 & 0.26  \\
5x3x5\_15\_3  & 140 & 60 & 0.22 & 0.81   & 0.42 & 0.72   & 0.04 & 0.31  \\
5x5x15\_20\_1 & 161 & 39 & 1.61 & 0.94   & 3.32 & 3.34   & 0.14 & 0.19  \\
5x5x15\_20\_2 & 192 & 8  & 1.44 & 0.88   & 3.53 & 2.88   & 0.14 & 0.34  \\
5x5x15\_20\_3 & 182 & 18 & 1.46 & 0.79   & 3.59 & 4.08   & 0.14 & 0.15  \\
5x5x10\_30\_1 & 185 & 15 & 3.29 & 324.75 & 4.72 & 234.26 & 0.34 & 39.43 \\
5x5x10\_30\_2 & 180 & 20 & 4.09 & 432.66 & 5.42 & 181.44 & 0.36 & 29.54 \\
5x5x10\_30\_3 & 199 & 1  & 2.91 & 43.75  & 5.23 & 38.04  & 0.30 & 16.33 \\
\hline%\multicolumn{9}{@{}c@{}}{}\\[-6.5pt]
\hline
Total       & 2030 & 370 & 1.39 & 37.02 & 2.45 & 20.26 & 0.14 & 3.36 \\

%% file: discussion.tex
\section{Discussion}\label{sec:discussion}

We have devised novel modeling approaches for continuous stream reasoning
based on ASP, utilizing time-decaying logic programs to capture
sliding window data in a natural way.
While such data is transient and subject to routine expiration,
we provided techniques to encode knowledge about potential window contents statically.
This reduces the dynamic tasks of re-grounding and integrating rules
of an offline encoding in view of new data
to matching inputs to a corresponding internal representation.
As a consequence, reasoning also concentrates on a fixed propositional
representation (whose parts are selectively activated wrt.\ actual window contents),
which enables the re-use of constraints gathered by  conflict-driven learning.
Although we illustrated modeling principles,
including an approach to propagate frame axioms along time slot cycles,
on toy domains only,
the basic ideas are of general applicability.
This offers interesting prospects for 
implementing knowledge-intense forms of stream reasoning,
as % needed for instance in 
required in application areas like
ambient assisted living, robotics, or dynamic scheduling.

Our approach to ASP-based stream reasoning is prototypically implemented
as an extension of the reactive ASP system \oclingo,
and preliminary experiments clearly show the need of improved  low-level
support of data expiration.
In fact,
we plan to combine the process of redesigning \oclingo\ with the addition
of yet missing functionalities, such as optimization in incremental and reactive settings.
Future work also regards the consolidation of existing and the addition of further
directives to steer incremental grounding and solving.
For instance,
beyond step-wise \code{\#cumulative} and \code{\#volatile} directives,
we envisage \code{\#assert} and \code{\#retract} statements,
as offered by Prolog (cf.~\cite{kowalski79a}), to selectively (de)activate logic program parts.
As with traditional ASP methods,
the objective of future extensions is to combine high-level declarative modeling with
powerful reasoning technology, automating   both grounding and search.
The investigation of sliding window scenarios performed here provides a first
step towards gearing ASP to continuous reasoning tasks.
Presupposing appropriate technology    support, we think   that many      dynamic domains
may   benefit from % powerful 
                            ASP-based reasoning.

%% file: acknowledgments.tex
\paragraph{Acknowledgments}
We are grateful to 
the anonymous referees
for helpful comments.
This work was partially funded 
by 
the German Science Foundation (DFG) 
under 
grant SCHA 550/8-1/2
and
by
the European Commission within % the EasyReach project
EasyReach
(\url{www.easyreach-project.eu})
under
grant AAL-2009-2-117.
% AAL call 2009-2.
% Federal Ministry of Education and Research within the GoFORSYS project
% (\url{http://www.goforsys.de}; grant~0313924).

%%% Local Variables: 
%%% mode: latex
%%% TeX-master: "paper"
%%% End: 

%% file: bbl.tex
%%% Local Variables: 
%%% mode: latex
%%% TeX-master: "paper"
%%% End: 